\documentclass{article}

\usepackage{arxiv}

\usepackage[utf8]{inputenc} 
\usepackage[T1]{fontenc}    
\usepackage{hyperref}       
\usepackage{url}            
\usepackage{booktabs}       
\usepackage{amsfonts}       
\usepackage{nicefrac}       
\usepackage{microtype}      
\usepackage{xcolor}         
\usepackage{natbib}
\usepackage{amsmath}
\usepackage{amssymb}
\usepackage{mathtools}
\usepackage{amsthm}
\usepackage{bm}
\usepackage{enumitem}
\usepackage{algorithm}
\usepackage{algorithmic}
\usepackage{multirow}
\usepackage{makecell}
\usepackage{wrapfig}
\usepackage{colortbl}
\usepackage{tcolorbox}
\tcbuselibrary{breakable}

\usepackage[capitalize,noabbrev]{cleveref}

\theoremstyle{plain}

\theoremstyle{definition}

\theoremstyle{remark}

\newcommand{\ourmethod}{EpiAgent}
\newcommand{\GrayComment}[1]{\STATE \textcolor{gray}{\bfseries // #1}}
\newcommand{\meanstd}[2]{\ensuremath{#1{\scriptstyle\,\pm\,#2}}}

\title{\ourmethod{}: Epistemic Self-Play for LLM Agents}

\author{
\textbf{Xuanfa Jin}$^{1,2}$ \enspace
\textbf{Xinyu Cui}$^{1,2}$ \enspace 
\textbf{Zhijian Ma}$^{1,2}$ \enspace
\textbf{Xue Yan}$^{1,2}$ \enspace
\textbf{Jian Zhao}$^{3}$ \\
\textbf{Haifeng Zhang}$^{1,2}$ \enspace
\textbf{Jun Wang}\thanks{Correspondence: \texttt{jun.wang@cs.ucl.ac.uk}}$^{~\,,4}$ \\ \\
$^1$Institute of Automation, Chinese Academy of Sciences \enspace
$^2$School of Artificial Intelligence, UCAS \\
$^3$Zhongguancun Academy \enspace
$^4$AI Centre, Department of Computer Science, UCL
}

\date{}

\renewcommand{\undertitle}{}
\renewcommand{\shorttitle}{}

\hypersetup{
pdftitle={\ourmethod{}: Epistemic Self-Play for LLM Agents},
pdfsubject={cs.AI, cs.LG},
pdfauthor={Xuanfa Jin, Xinyu Cui, Zhijian Ma, Xue Yan, Jian Zhao, Haifeng Zhang, Jun Wang},
pdfkeywords={LLM-based Agents, Multi-Agent Interactions, Epistemic Reasoning, Self-Play},
}

\begin{document}

\maketitle

\begin{abstract}
Epistemic game theory explains an agent's strategic behavior via the beliefs it holds about others and how these beliefs are updated as the interactions unfold. Modern LLM-based agents, however, often lack an explicit mechanism for learning and updating such beliefs to guide decisions under incomplete information. To bridge this gap, we propose \textbf{\ourmethod{}}, a framework that equips LLM-based agents with an explicit, learnable epistemic layer, which maintains beliefs over others' private information and combines these beliefs with the agent's available information into an epistemic state for action selection. Concretely, the agent's policy is decomposed into two coupled stages: \textit{Epistemic Update}, which updates these beliefs from interaction histories, and \textit{Action Planning}, which selects actions conditioned on the updated epistemic state. By formulating their dependency as a bi-level optimization problem, we introduce \textbf{Epistemic Self-Play}, an algorithm that alternately optimizes the two stages through self-play, utilizing an intrinsic prediction reward to evaluate how well inferred beliefs predict subsequent observations. Empirical results across both mixed and competitive scenarios show that \ourmethod{} consistently outperforms prompting-based and RL-based baselines. Further analyses reveal the benefits of dedicated epistemic training, improvements in belief inference accuracy, and the contribution of learned beliefs to action planning. These results demonstrate the value of an explicit, learnable epistemic layer for effective multi-agent interactions.
\end{abstract}

\section{Introduction}
Large Language Models (LLMs) have demonstrated impressive capabilities in reasoning and planning \citep{achiam2023gpt, guo2025deepseek}, sparking a growing interest in deploying LLM-based agents with Multi-Agent Systems (MAS) to tackle complex tasks \citep{qian2024chatdev, liang2024encouraging, schmidgall2025agent}. Unlike single-agent tasks where LLMs have excelled \citep{lu2025uxagent, jin2025search}, multi-agent scenarios introduce significant non-stationarity and incomplete information. In these settings, environment dynamics and outcomes are determined not only by the ego agent's own decisions but also by the policies of other agents, who often possess private information that is unobservable to others \citep{littman1994markov, albrecht2024multi}. Since such private information can influence other agents' behavior, inferring it from interaction histories can help an agent anticipate their future behaviors and adapt its own decisions accordingly.

Inspired by epistemic game theory (EGT) \citep{aumann1999interactive, dekel2015epistemic}, we view an agent's strategic behavior as being explained via the beliefs that agents hold about others, and the way these beliefs are updated through the game. From this perspective, an agent's decision can be naturally understood in terms of two coupled components: an \textit{interim belief} over the private information of others, updated from the observed interaction history, and a \textit{belief-conditioned action} that is guided by this belief. The agent's own private information, observed history and its current belief jointly constitute its epistemic state, which guides the agent's strategic reasoning and action selection \citep{perea2012epistemic}. Many LLM-based agents, however, lack such an explicit epistemic layer in their decision-making pipeline. They usually map observations directly to actions \citep{nakano2021webgpt, wang2023voyager}, or rely on prompting to elicit free-form reasoning \citep{wei2022chain, zhao2024expel} that is treated as a byproduct of generation rather than as a structured epistemic process. Even when explicit opponent modeling is introduced, it is typically decoupled from the agent's policy learning and does not co-evolve with it \citep{cross2024hypothetical, kim2025hypothesis}. This motivates developing an explicit, learnable epistemic layer whose learning is coordinated with that of the action policy through interaction.

To this end, we propose \ourmethod{}, a framework that equips LLM-based agents with an explicit epistemic layer for decision-making. We first formulate multi-agent interaction scenarios with fixed private information within a partially observable stochastic game (POSG) \citep{hansen2004dynamic} framework, which formally specifies each agent's private information and the interaction dynamics of multiple agents. Within this framework, an agent's policy is unfolded into two coupled stages: \textit{Epistemic Update}, which updates the agent's belief from its observed interaction histories, and \textit{Action Planning}, which selects actions based on the updated epistemic state. However, training such an agent is non-trivial due to the mutual dependence between the two stages: epistemic update learns from interaction trajectories generated by the action policy, while action planning learns to make decisions based on the evolving beliefs. We cast this dependency as a bi-level optimization problem and introduce Epistemic Self-Play, an algorithm that alternately optimizes the two stages through self-play. Given that the ground-truth labels for other agents' private information are often inaccessible in real-world scenarios, we further introduce an intrinsic prediction reward to train epistemic updates with subsequent observations.

To validate the performance of our method, we adopt two distinct yet challenging multi-agent environments: \textit{Bargaining} and \textit{Kuhn Poker}, both of which exemplify the complexities of incomplete information. The \textit{Bargaining} scenario represents a mixed negotiation game where agents must trade off self-interest against the risk of breakdown. Meanwhile, the \textit{Kuhn Poker} scenario represents a competitive game involving stochasticity and deception. Empirical results demonstrate that \ourmethod{} greatly outperforms both prompting-based baselines and an RL-trained baseline without an explicit epistemic layer. Furthermore, ablation results demonstrate that dedicated training of the epistemic update stage provides additional performance benefits, while evaluations of belief inference show improved accuracy over training. Belief interventions further show that the learned beliefs contribute to effective action planning. These results suggest that the performance gains of our method are not solely attributable to action planning optimization, but also benefit from improved epistemic reasoning. Our main contributions are summarized as follows: 
\begin{itemize}[leftmargin=12pt, topsep=3pt, itemsep=0pt]
    \item We propose \textbf{\ourmethod{}}, an LLM-based agent framework that introduces an explicit epistemic layer into the agent's decision-making pipeline by decomposing the agent's policy into an \textit{Epistemic Update} stage and an \textit{Action Planning} stage.
    
    \item We introduce \textbf{Epistemic Self-Play}, an algorithm designed to promote the co-evolution of epistemic update and action planning through alternating self-play, equipped with an intrinsic prediction reward that trains the agent's epistemic inference without ground-truth labels.
    
    \item We provide comprehensive empirical results showing that \ourmethod{} outperforms both prompting-based and RL-based baselines, and that its learned beliefs effectively improve action planning, demonstrating the value of an explicit, learnable epistemic layer for multi-agent interactions.
\end{itemize}

\section{Related Work}
\paragraph{LLM-based Agent for Decision-Making.}
Recent advancements in Large Language Models (LLMs) have significantly expanded their utility from natural language processing to autonomous decision-making tasks \citep{zhu2023ghost, he2024webvoyager, zhang2025socioverse}. Foundational techniques such as Chain-of-Thought (CoT) \citep{wei2022chain} enable LLMs to decompose complex tasks into simpler ones, while frameworks like ReAct \citep{yao2022react} and Reflexion \citep{shinn2023reflexion} further enhance agents' decision-making capabilities by interleaving reasoning with action execution. Beyond inference-time prompting, recent works have also explored Reinforcement Learning (RL) to fundamentally improve agent policies \citep{zhang2025landscape}, particularly for tasks with verifiable rewards, such as mathematics \citep{shao2024deepseekmath, zuo2025ttrl} and coding \citep{wei2025swe, zeng2025acecoder}. However, extending these successes to multi-turn scenarios is non-trivial, as the immediate ground-truth feedback in those scenarios is often sparse or absent. To address this, SWEET-RL \citep{zhou2025sweet} adopts a turn-level critic to improve credit assignments, thereby accelerating the learning process. And RAGEN \citep{wang2025ragen} provides a unified framework for multi-turn, trajectory-level LLM-based agent training. While these methods can be directly applied to multi-agent scenarios, they do not explicitly model other agents' private information. By equipping the LLM-based agent with an explicit epistemic layer, \ourmethod{} can infer such information from interaction histories to guide its subsequent actions.

\begin{figure*}[t]
  \centering
  \vspace{-0.4cm}
  \includegraphics[width=\linewidth]{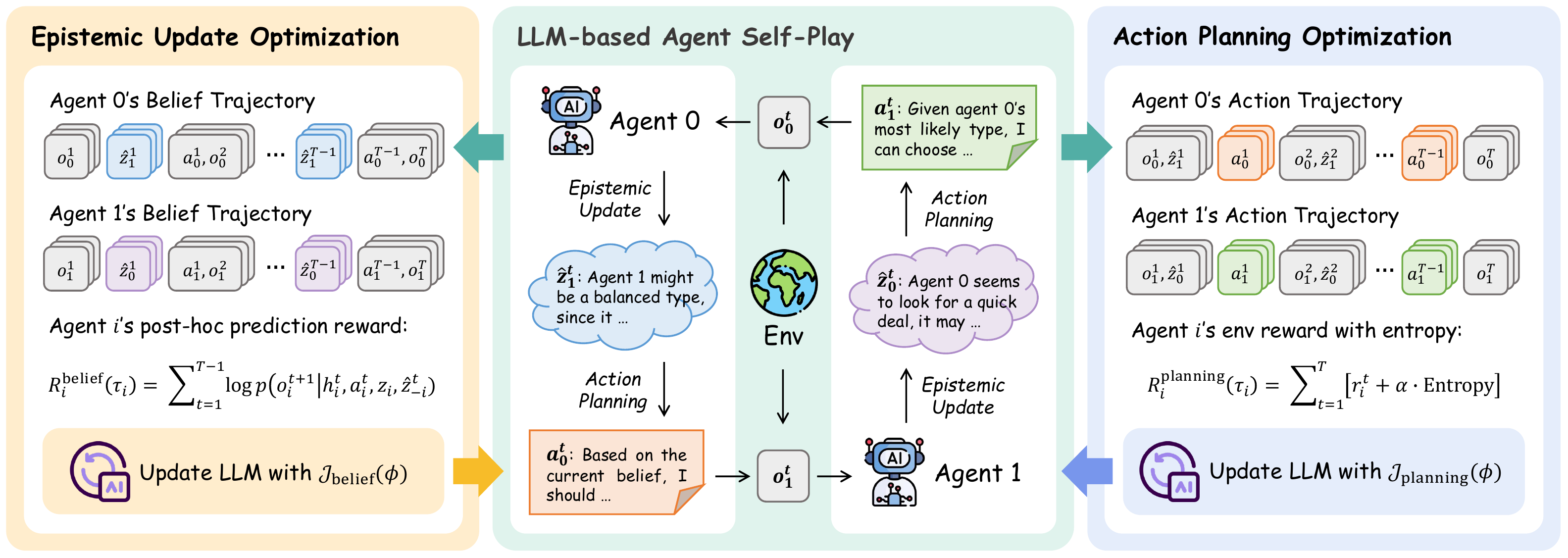}
  \vspace{-0.5cm}
  \caption{The \ourmethod{} framework trained with Epistemic Self-Play. Agent 0 and Agent 1 share a single LLM, performing self-play in the multi-agent environment and collecting interaction histories (middle). The LLM is then optimized alternately between the epistemic update (left) and the action planning (right) stage, by reconstructing the histories into corresponding trajectories for RL training.}
  \label{fig:framework}
  \vspace{-0.4cm}
\end{figure*}

\paragraph{LLM-based Multi-Agent Systems.}
Applying LLMs in multi-agent systems (MAS) is becoming a vibrant research area, and the applications can be broadly categorized into world simulation and problem solving \citep{li2024survey}. In world simulation, Generative Agents \citep{park2023generative} construct a sandbox environment to observe human-like behaviors emerge from LLM-based agents. Furthermore, strategy games such as Diplomacy \citep{guan2024richelieu, peng2025diplomacyagent}, Werewolf \citep{xu2024language, jin2024learning}, and Avalon \citep{shi2023cooperation, lan2024llm} are often adopted to test LLMs' strategic decision-making abilities. When it comes to problem solving, researchers mainly focus on leveraging role-playing and collaboration to tackle various tasks. CAMEL \citep{li2023camel} and AutoGen \citep{wu2024autogen} design cooperative agents for general reasoning and question answering. SWE-Agent \citep{yang2024swe} and ChatDev \citep{qian2024chatdev} incorporate standard operating procedures (SOPs) into agent interactions, simulating human workflows where agents with specialized roles collaborate to develop software. Conversely, Multi-Agent Debate \citep{liang2024encouraging, liu2025breaking, jin2026remember} takes a competitive way by encouraging agents to critique each other's views to reach a consensus, thereby improving the quality of the final solution. These works mostly focus on cooperative settings or reason about other agents implicitly, without explicitly modeling their private information. In contrast, we integrate and train an epistemic update stage to explicitly infer others' private information, enabling agents' adaptive decision-making across various scenarios.

\paragraph{Belief Modeling and Theory of Mind in LLM Agents.}
A growing body of work has investigated whether and how LLMs can reason about others' mental states. Many works study LLMs' Theory of Mind (ToM) \citep{frith2005theory} capabilities through false-belief tasks and social reasoning benchmarks \citep{gandhi2023understanding, wu2023hi, shinoda2025tomato}, or attempt to enhance such capabilities through prompt engineering \citep{li2023theory}, workflow design \citep{zhang2025autotom, kim2025hypothesis}, or fine-tuning \citep{lu2025theory, hwang2025infusing}. However, these methods mostly treat ToM as a standalone capability to be evaluated or improved, rather than as an active component in the agent's decision-making pipeline. Epistemic game theory provides a perspective on this connection, explaining strategic behavior through the beliefs agents hold about others and how these beliefs are updated as the interaction unfolds \citep{aumann1999interactive, dekel2015epistemic, perea2012epistemic}. Some works have begun to use such reasoning for decision-making: Hypothetical Minds \citep{cross2024hypothetical} prompts LLMs to maintain hypotheses about other agents' strategies, which further guide action generation, and \citet{zhang2024mutual} conduct an empirical study of the Mutual ToM process in human-AI collaboration. However, these methods mainly rely on the inference-time capabilities of a fixed LLM, leaving belief inference decoupled from the agent's policy learning. In multi-agent RL, Policy Belief Learning \citep{tian2020learning} alternately trains a belief module and a belief-conditioned policy, but supervises beliefs with ground-truth private information. In contrast, \ourmethod{} trains epistemic update and action planning within the same LLM through self-play, learning to infer others' private information from interaction feedback without ground-truth labels and to act on the resulting beliefs.

\section{Preliminaries}
\label{sec:formulation}

\paragraph{Problem Formulation.}
We consider finite-horizon, turn-based interactions among agents with private information. Following the static private-information setting of \citet{tian2020learning}, we distinguish each agent's fixed private information from the dynamic interaction state. We describe the environment by \(\mathcal{E}=\langle \mathcal{N},\allowbreak \mathcal{X},\allowbreak \mathcal{Z},\allowbreak \mathcal{A},\allowbreak \mathcal{O},\allowbreak \rho,\allowbreak P,\allowbreak \{r_i\}_{i\in\mathcal{N}}\rangle\), where \(\mathcal{N}=\{1,\ldots,n\}\) is the agent set and \(\mathcal{X}\) is the space of dynamic states. The joint private-information, action, and observation spaces are denoted as \(\mathcal{Z}=\times_i\mathcal{Z}_i\), \(\mathcal{A}=\times_i\mathcal{A}_i\), and \(\mathcal{O}=\times_i\mathcal{O}_i\), respectively.

At the start of an episode, \((x^1,\mathbf{z},\mathbf{o}^1) \sim \rho\) specifies the initial state, each agent's private information, and their observations. Each \(z_i\in\mathcal{Z}_i\) is available to agent \(i\) but not directly observable by other agents, and remains fixed throughout the episode. At interaction step \(t\), the joint action \(\mathbf{a}^t\) induces \((x^{t+1},\mathbf{o}^{t+1})\sim P(\cdot\mid x^t,\mathbf{z},\mathbf{a}^t)\), where \(P\) is the transition and observation kernel, and agent \(i\) receives reward \(r_i^t=r_i(x^t,\mathbf{z},\mathbf{a}^t)\). The dynamic state \(x^t\) may itself be partially observed. Turn-taking is encoded in \(x^t\), with inactive agents restricted to a null action. This formulation is an instance of a partially observable stochastic game (POSG) \citep{hansen2004dynamic} with full state \(s^t=(x^t,\mathbf{z})\), whose static component \(\mathbf{z}\) is preserved by the transition.
Agent \(i\)'s interaction history is denoted as \(h_i^t=(o_i^1,\allowbreak a_i^1,\allowbreak \ldots,\allowbreak a_i^{t-1},\allowbreak o_i^t)\), where \(o_i^t\) is the observation available to that agent at step \(t\), including other agents' actions when observable. Together, the interaction history \(h_i^t\) and the agent's own private information \(z_i\) constitute the information available to agent \(i\) for decision-making. Thus, agent \(i\)'s policy can be defined as \(\pi_i(\cdot \mid h_i^t,z_i)\).

\paragraph{Beliefs and Epistemic States.}
In partially observable decision problems, a \textit{belief} is a posterior distribution over the underlying state given an agent's available information \citep{kaelbling1998planning}. In our setting, we focus on the agent's belief over others' fixed private information, which is inferred based on its interaction history. Specifically, agent \(i\)'s belief at step \(t\) is defined as \(b_i^t \in \Delta(\mathcal{Z}_{-i})\), where \(\mathcal{Z}_{-i} = \times_{j \neq i} \mathcal{Z}_j\). Although \(z_{-i}\) remains fixed within an episode, its posterior can be updated as new observations arrive. Together with the agent's own private information and interaction history, this belief defines agent \(i\)'s \textit{epistemic state} at each step:
\begin{equation}
\label{eq:epistemic_state}
    e_i^t=(z_i,h_i^t,b_i^t), \quad b_i^t \in \Delta(\mathcal{Z}_{-i}),
\end{equation}
where \(z_i\) and \(h_i^t\) summarize what agent \(i\) knows, and \(b_i^t\) summarizes what it infers about others. While epistemic game theory in its full generality concerns belief hierarchies of arbitrary order \citep{dekel2015epistemic}, here we focus on the first-order layer, each agent's belief over others' private information, as the foundational case for LLM-based agents, leaving extensions to higher-order beliefs for future work.

\paragraph{Policy Optimization.}
In reinforcement learning, an agent's goal is usually to learn a policy that maximizes its expected cumulative reward. Assuming the agent's policy \(\pi\) is parameterized by \(\phi\), the optimization objective is typically formulated as:
\begin{equation}
\label{eq:interaction_objective}
    \mathcal{J}(\phi) = \mathbb{E}_{\tau \sim \pi_\phi} \left[ \sum_{t=1}^{T}r^t \right],
\end{equation}
where \(\tau\sim\pi_\phi\) denotes a trajectory generated under \(\pi_\phi\), and \(T\) denotes the episode length. To optimize this objective, policy gradient (PG) methods \citep{sutton1999policy} are widely adopted to update the policy parameters. The gradient of the objective is given by:
\begin{equation}
\label{eq:policy_gradient}
    \nabla_\phi \mathcal{J}(\phi) = \mathbb{E}_{\tau\sim\pi_\phi}\left[
        \sum_{t=1}^{T} \nabla_\phi\log\pi_\phi(a^t\mid h^t) \cdot G^t
    \right],
\end{equation}
where \(h^t\) denotes the information available to the policy before taking action \(a^t\), and \(G^t\) denotes the return signal at step \(t\), instantiated as cumulative reward, return-to-go, advantage, or other variants depending on the specific algorithms.

\section{\ourmethod{}: Framework and Training}
\label{sec:method}

\begin{wrapfigure}{r}{0.52\textwidth}
    \centering
    \vspace{-0.8cm}
    \includegraphics[width=0.9\linewidth]{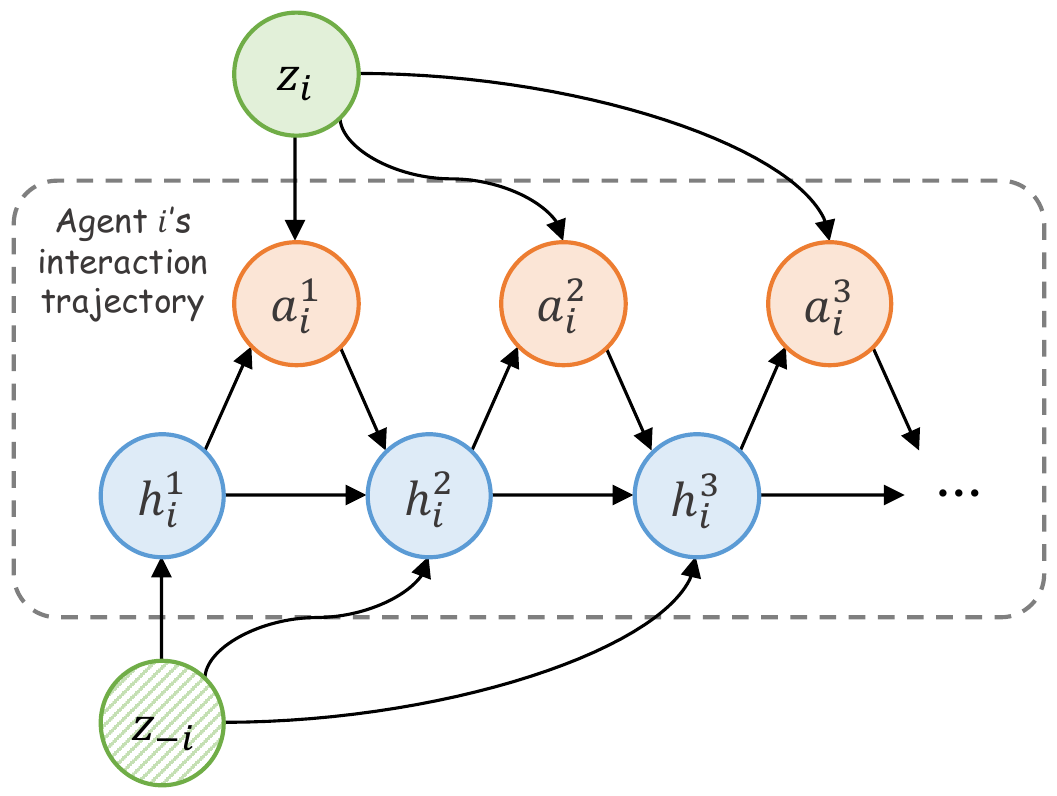}
    \vspace{-0.1cm}
    \caption{The generative model of multi-agent interaction from agent \(i\)'s perspective. Agent \(i\) makes decisions based on its own private information \(z_i\) and the observed history \(h_i^t\), while the transition of \(h_i^t\) is driven by other agents' actions, which depend on their private information \(z_{-i}\).}
    \label{fig:generative_model}
    \vspace{-1.2cm}
\end{wrapfigure}

In this section, we present \textbf{\ourmethod{}}, a framework that equips LLM-based agents with an explicit epistemic layer for decision-making, as shown in \cref{fig:framework}. We first formalize the decomposition of an agent's policy into two coupled stages, \textit{Epistemic Update} and \textit{Action Planning}, and cast their joint training as a bi-level optimization problem. We then develop optimization objectives for both stages, including an intrinsic prediction reward that enables learning epistemic updates from interaction feedback without ground-truth supervision of other agents' private information. Finally, we introduce \textbf{Epistemic Self-Play}, an alternating training algorithm designed to promote the co-evolution of epistemic update and action planning through self-play.

\subsection{Epistemic Decomposition of Agent Policy}
\label{sec:epistemic_decomposition}
To formalize how agents make decisions in multi-agent interactions, we begin by modeling the interaction process from a single agent's perspective. The generative model in \cref{fig:generative_model} describes the evolution of agent \(i\)'s interaction history. At step \(t\), agent \(i\) selects action \(a_i^t\) based on its own private information \(z_i\) and history \(h_i^t\). The subsequent observation \(o_i^{t+1}\) extends this history as \(h_i^{t+1}=(h_i^t,a_i^t,o_i^{t+1})\). Other agents' behavior can depend on their fixed private information \(z_{-i}\) and interaction histories, making the observed interaction a source of evidence for inferring \(z_{-i}\).

From an epistemic perspective, an agent's decision-making can be understood in terms of what it knows about the interaction and what it believes about others. These components are represented by its epistemic state \(e_i^t=(z_i,h_i^t,b_i^t)\). Building on this perspective, we unfold agent \(i\)'s policy \(\pi_i\) into two coupled stages:
\begin{itemize}[leftmargin=12pt, topsep=1pt, itemsep=1pt]
    \item \textit{Epistemic Update}: Given its interaction history \(h_i^t\) and own private information \(z_i\), agent \(i\) first maintains a belief model \(b_i\) to update its current belief \(b_i^t = b_i(\cdot\mid h_i^t,z_i)\). As new observations become available, this model revises the agent's belief about the fixed private information \(z_{-i}\).

    \item \textit{Action Planning}: Given a hypothesis \(\hat z_{-i}^t\) about other agents' private information, agent \(i\) then employs a planning model \(\mu_i(\cdot\mid h_i^t,z_i,\hat z_{-i}^t)\) to generate an action \(a_i^t\). The planning model combines this hypothesis with the interaction history and the agent's own private information.
\end{itemize}
Together, these two stages define a decomposition of the agent's overall policy: at each step, the action distribution is obtained by marginalizing the planning model over its updated belief:
\begin{equation}
\label{eq:decomposition}
    \pi_i(a_i^t\mid h_i^t,z_i)
    =\mathbb{E}_{\hat z_{-i}^t\sim b_i(\cdot \mid h_i^t, z_i)}
    \left[\mu_i(a_i^t\mid h_i^t,z_i,\hat z_{-i}^t)\right].
\end{equation}
This policy can be realized by first sampling \(\hat z_{-i}^t\sim b_i(\cdot \mid h_i^t, z_i)\) and then generating \(a_i^t\sim\mu_i(\cdot\mid h_i^t,z_i,\hat z_{-i}^t)\). In this way, inference about other agents becomes an explicit intermediate step in the decision process. This sequential generation process naturally aligns with the autoregressive structure of LLM-based agents.

The decomposition gives rise to two interdependent learning objectives: action planning learns to make decisions under the inferred beliefs, while epistemic update draws on evidence from the interactions generated by the planning model. We formalize their joint training as a bi-level optimization problem:
\begin{equation}
\label{eq:bilevel_formulation}
    \begin{aligned}
        \max_{\mu_i}\quad &
        \mathcal{J}_{\mathrm{planning}}\bigl(\mu_i,b_i^\star(\mu_i)\bigr),\\
        \text{s.t.}\quad &
        b_i^\star(\mu_i)\in\arg\max_{b_i}
        \mathcal{J}_{\mathrm{belief}}(b_i,\mu_i).
    \end{aligned}
\end{equation}
Here, \(\mathcal{J}_{\mathrm{belief}}\) and \(\mathcal{J}_{\mathrm{planning}}\) are the objectives of epistemic update and action planning, respectively. For a given planning model \(\mu_i\), the inner problem optimizes the belief model using the induced trajectories. The outer problem evaluates the planning model together with its corresponding adapted belief model. We develop these objectives in \cref{sec:optimization_objectives} and introduce Epistemic Self-Play in \cref{sec:epistemic_self_play} to train the two coupled stages.

\subsection{Optimization Objectives}
\label{sec:optimization_objectives}

\paragraph{Epistemic Update.}
The goal of the epistemic update stage is to infer other agents' private information \(z_{-i}\) from the interaction history. However, a primary challenge in realistic multi-agent interaction scenarios is that the private information is usually unrevealed, making direct supervision of belief inference difficult. We address this challenge by learning from how well the inferred private information predicts subsequent observations. Following Bayesian inverse planning \citep{baker2009action, zhi2020online}, we motivate this learning signal through the posterior:
\begin{equation}
\label{eq:bayes}
    p(z_{-i}\mid h_i^t,z_i) \propto p(z_{-i}\mid z_i) \cdot \prod_{k=1}^{t} p(o_i^k\mid h_i^{k-1},a_i^{k-1},z_i,z_{-i}).
\end{equation}
Here, \(p(z_{-i}\mid z_i)\) is the prior conditioned on the agent's own private information. We set \(h_i^0 = a_i^0 = \varnothing\) as placeholders before the initial observation, so that the \(k=1\) factor denotes \(p(o_i^1\mid z_i,z_{-i})\). Each likelihood factor measures how well a candidate \(z_{-i}\) explains the observed interactions. This formulation suggests a criterion for evaluating epistemic inference: although other agents' private information is not directly observed, an inferred hypothesis can be assessed by how well it predicts their observed behavior. Interaction feedback can therefore provide a learning signal for the belief model without requiring ground-truth labels.

Based on this insight, we introduce an intrinsic prediction reward to train the epistemic update stage. Specifically, after sampling \(\hat{z}_{-i}^t \sim b_i(\cdot \mid h_i^t, z_i)\) and taking action \(a_i^t\), the agent receives a new observation \(o_i^{t+1}\), which provides feedback on the sampled hypothesis. We compute the post-hoc reward as the log-likelihood of this observation under that hypothesis:
\begin{equation}
\label{eq:pred_reward}
    r_{i,\mathrm{pred}}^t(\hat{z}_{-i}^t) = \log p(o_i^{t+1}\mid h_i^t,a_i^t,z_i,\hat z_{-i}^t).
\end{equation}
In practice, this likelihood is computed by the agent's own LLM with a dedicated prediction prompt (\cref{appendix:prompt}), normalized by the length of \(o_i^{t+1}\). Since other agents' responses are generated by the same LLM during self-play, it naturally serves as an approximate model of their behavior.
This reward encourages the belief model to produce hypotheses under which the actual subsequent observations are more likely to occur. Accordingly, the objective for the agent \(i\)'s epistemic update stage is to maximize the expected cumulative intrinsic reward:
\begin{equation}
\label{eq:belief_obj}
    \mathcal{J}_{\mathrm{belief}}(b_i,\mu_i)
    =\mathbb{E}_{b_i,\mu_i}
    \left[\sum_{t=1}^{T-1}r_{i,\mathrm{pred}}^t(\hat{z}_{-i}^t)\right],
\end{equation}
where \(T\) denotes the episode length. By optimizing this objective, the belief model is optimized purely by the agent's own predictive consistency, bypassing the need for ground-truth labels of other agents' private information.

\paragraph{Action Planning.}
Given a hypothesis \(\hat z_{-i}^t\sim b_i(\cdot\mid h_i^t,z_i)\) sampled during epistemic update, agent \(i\)'s planning model \(\mu_i\) selects actions conditioned on this hypothesis, its interaction history \(h_i^t\), and its own private information \(z_i\). With the latent variables that influence the interaction dynamics explicitly estimated as \(\hat z_{-i}^t\), the planning model uses these estimates to guide action selection, with the primary objective of maximizing the expected cumulative environment reward. However, the beliefs used for planning are themselves learned and continue to evolve during training. Consequently, the sampled hypothesis \(\hat z_{-i}^t\) may not accurately reflect other agents' private information. Concentrating too early on a narrow set of actions under such estimates may limit the exploration of alternative policies while the belief model is still adapting. We therefore encourage the agent to maintain diversity in its action selection conditioned on the inferred beliefs, allowing broader exploration of potential policies as its beliefs evolve.

To this end, we adopt a maximum-entropy reinforcement learning framework that encourages exploration over belief-conditioned policies. In addition to maximizing the cumulative reward, this framework augments the reward with the entropy of the conditional action distribution. Accordingly, the optimization objective for agent \(i\)'s action planning stage is defined as:
\begin{equation}
\label{eq:planning_obj}
    \mathcal{J}_{\mathrm{planning}}(\mu_i,b_i)
    =\mathbb{E}_{b_i,\mu_i}
    \left[\sum_{t=1}^{T}
    \left(r_i^t+\alpha\mathcal{H}(\mu_i^t)\right)\right],
\end{equation}
where \(\mu_i^t\) abbreviates the action distribution \(\mu_i(\cdot\mid h_i^t,z_i,\hat z_{-i}^t)\), \(\mathcal{H}(\cdot)\) denotes the entropy, and \(\alpha \geq 0\) is a coefficient that controls the trade-off between reward and action diversity.

\begin{algorithm}[!ht]
\caption{Epistemic Self-Play}
\label{alg:training}
\begin{algorithmic}[1]
    \REQUIRE Pretrained LLM \(p_\phi\), switch frequency \(L\)
    \GrayComment{\(0\) for Epistemic Update, \(1\) for Action Planning}
    \STATE Set initial training stage \(\psi\gets 0\)
    \WHILE{not done}
        \FOR{\(l=1\) to \(L\)}
            \STATE Rollout \(M\) self-play episodes using \(p_\phi\)
            \FOR{\(i=1\) to \(n\)}
                \STATE Extract agent \(i\)'s trajectories \(\mathcal{B}_i = \{ \tau_i^{(m)} \}_{m=1}^M\), where \(\tau_i^{(m)}=\left(o_i^{(m),t}, \hat{z}_{-i}^{(m),t}, a_i^{(m),t}\right)_{t=1}^T\)
                \STATE Calculate role-based advantages by Eq.~\eqref{eq:advantage}
            \ENDFOR
            \STATE Combine all agents' trajectories \(\mathcal{B}=\bigcup_{i=1}^n\mathcal{B}_i\)
            \STATE Update \(p_\phi\) on \(\mathcal{B}\) using policy gradient with objective Eq.~\eqref{eq:belief_obj} (\(\psi=0\)) or Eq.~\eqref{eq:planning_obj} (\(\psi=1\))
        \ENDFOR
        \STATE Switch training stage \(\psi\gets 1-\psi\)
    \ENDWHILE
\end{algorithmic}
\end{algorithm}

\subsection{Epistemic Self-Play}
\label{sec:epistemic_self_play}
Training the two-stage decision process involves a hierarchical dependency: the action planning stage requires stable and accurate beliefs to optimize its policy, while the epistemic update stage requires informative interaction trajectories generated by the action policy to learn different patterns in other agents' behavior. To coordinate learning under this dependency, we introduce \textbf{Epistemic Self-Play}, an alternating training algorithm designed to promote the co-evolution of epistemic update and action planning. The algorithm alternates between optimizing the two stages using interaction trajectories collected through self-play, allowing each stage to learn from feedback shaped by the evolving other stage. Meanwhile, self-play provides experience against continually updated interaction partners, with the aim of reducing specialization to a single fixed training opponent.

To realize this framework, we seek a unified agent that can both infer other agents' private information and use these inferences to guide its own actions. Therefore, we instantiate both the belief model \(b_i\) and the planning model \(\mu_i\) using a single LLM parameterized by \(\phi\). Specifically, we adopt \(p_\phi(\hat z_{-i}^t\mid h_i^t,z_i)\) to represent agent \(i\)'s belief model, and \(p_\phi(a_i^t\mid h_i^t,z_i,\hat z_{-i}^t)\) to represent its planning model. This implementation enables parameter sharing across the two stages. At each training iteration, the current model \(p_\phi\) engages in self-play, acting as all agents to generate complete episodes. We then extract each agent's local interaction trajectories and combine them into a training batch \(\mathcal B\). Using this batch, we update \(\phi\) through policy gradient to optimize the objective of the current training stage, namely \(\mathcal{J}_{\mathrm{belief}}\) or \(\mathcal{J}_{\mathrm{planning}}\). Finally, after certain iterations, training switches to the other stage. This alternating schedule approximates the bi-level problem in Eq. \eqref{eq:bilevel_formulation}, allowing each stage to adapt to the other's changes. The complete training procedure is outlined in \cref{alg:training}.

A practical concern in multi-agent training is that different agents may exhibit asymmetric expected returns due to inherent role asymmetries. Directly combining raw trajectories from different agents can introduce high variance, destabilizing the training process. Inspired by \citet{liu2025spiral}, we adopt a role-based advantage estimation:
\begin{equation}
\label{eq:advantage}
    A_i^{\psi}(\tau_i)=R_i^{\psi}(\tau_i)-\operatorname{mean}(\mathbf{R}_i^{\psi}),
\end{equation}
where \(\psi\) indicates the stage, \(R_i^{\psi}(\tau_i)\) is agent \(i\)'s return for that stage on trajectory \(\tau_i\), and \(\mathbf{R}_i^{\psi}\) contains the corresponding returns across all rollout trajectories. This approach helps the unified LLM to learn a robust policy that generalizes across various roles.

\section{Experiments}
\label{sec:experiment}
In this section, we design experiments to investigate the effectiveness of our proposed \ourmethod{} framework and the Epistemic Self-Play training algorithm. Specifically, we focus on answering the following research questions:

\begin{itemize}[leftmargin=12pt, topsep=3pt, itemsep=0pt]
    \item \textbf{RQ1: How does \ourmethod{} perform in multi-agent interaction scenarios?}
    We investigate whether \ourmethod{}, which integrates explicit epistemic update with belief-conditioned action planning, can improve the agent's overall performance in multi-agent interactions.
    
    \item \textbf{RQ2: Does training the epistemic update provide additional performance benefits?}
    We investigate whether learning epistemic updates from interaction feedback helps improve action planning, where the agent makes decisions based on its evolving beliefs.
    
    \item \textbf{RQ3: Do learned beliefs provide useful information for action planning?}
    We examine how well beliefs learned from intrinsic prediction reward reflect opponents' private information, and whether these inferred beliefs help guide the agent's subsequent action planning.
\end{itemize}

\subsection{Experiment Setups} \label{sec:exp_setup}
We train and evaluate LLM-based agents in modified two-player versions of \textit{Bargaining} and \textit{Kuhn Poker}. In both environments, each agent, regardless of its method, is privately assigned by the environment a behavioral persona that remains fixed within an episode and serves as the main private information for epistemic inference. We introduce three personas in Bargaining and four in Kuhn Poker that encourage different negotiation and betting styles. Bargaining additionally includes private target and bottomline prices, while our modified Kuhn Poker comprises five rounds with newly dealt private cards in each round. Detailed environment settings are provided in \cref{appendix:env}.

In our experiments, we employ Qwen2.5-3B-Instruct as our base model for all methods and use LoRA for training. Models are trained based on REINFORCE++ \citep{hu2025reinforce} with role-based advantage estimation (Eq. \eqref{eq:advantage}) for approximately 120 rollout-update iterations. The switch frequency \(L\) for our Epistemic Self-Play is set to \(10\), and training starts from the epistemic update stage. During evaluation, model parameters remain fixed. We conduct 128 episodes per agent pairing in each evaluation run with a temperature of \(0.5\). Further training and evaluation details can be found in \cref{appendix:exp_details}.

We adopt the following baselines in our experiments: \textbf{ReAct} \citep{yao2022react} integrates reasoning into the decision-making process in LLMs, allowing models to generate thought traces before taking actions; \textbf{Reflexion} \citep{shinn2023reflexion} evaluates LLM-based agent's current states and actions through self-reflection, providing feedback to guide its subsequent decisions; and \textbf{Hypothetical Minds (HM)} \citep{cross2024hypothetical} enhances the LLM-based agent through a prompting-based opponent modeling module that generates and evaluates hypotheses about other agents' policies. Also, we employ \textbf{ReAct+RL} as a training baseline, which applies the same RL algorithm to fine-tune the ReAct agent through self-play with the same number of iterations, optimizing its policy directly from the environment rewards. Furthermore, we include \textbf{\ourmethod{} w/o Epistemic Training} as an ablation baseline, which retains both stages of the decision process but optimizes only the action planning stage.

\begin{figure*}[t]
    \centering
    \vspace{-0.3cm}
    \includegraphics[width=0.95\textwidth]{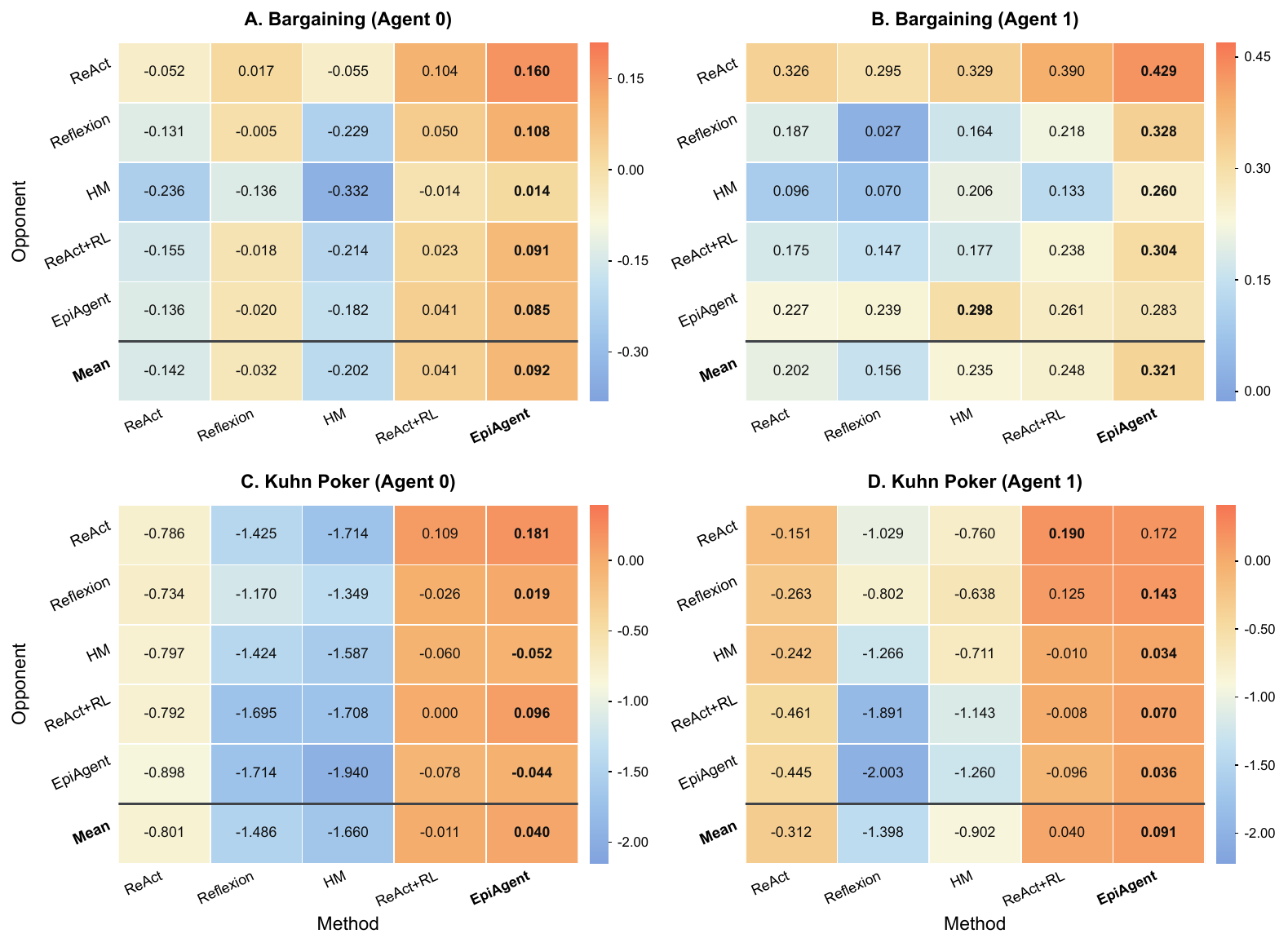}
    \vspace{-0.2cm}
    \caption{Cross-play evaluation results in \textit{Bargaining} and \textit{Kuhn Poker}, split by evaluated role. Columns identify the evaluated method and rows identify its opponent. Each cell reports the average reward (with format penalty) for the method. The bottom row averages over the five opponents. Full results with standard deviations are provided in \cref{tab:cross_play_results}.}
    \label{fig:crossplay_heatmap}
    \vspace{-0.2cm}
\end{figure*}

\begin{table}[t]
    \centering
    \caption{Action validity diagnostics for methods evaluated as Agent 0 in both environments. Valid reward is the mean reward over episodes in which neither agent commits a violation, averaged equally over opponents.}
    \label{tab:validity_diagnostics}
    \small
    \renewcommand{\arraystretch}{1.1}
    \begin{tabular}{lrrrr}
        \toprule
        \multirow{2}{*}{\textbf{Method}} & \multicolumn{2}{c}{\textit{Bargaining}} & \multicolumn{2}{c}{\textit{Kuhn Poker}} \\
        \cmidrule(lr){2-3}\cmidrule(lr){4-5}
        & Violation (\%) & Valid reward & Violation (\%) & Valid reward \\
        \midrule
        \textbf{ReAct}        & 0.00 & -0.142 & 22.54 & -0.189 \\
        \textbf{Reflexion}    & 0.36 & -0.021 & 47.84 & -0.111 \\
        \textbf{HM}           & 0.00 & -0.203 & 55.34 & -0.015 \\
        \textbf{ReAct+RL}     & 0.00 &  0.041 &  0.06 & -0.026 \\
        \rowcolor{cyan!15} \textbf{\ourmethod{}} & 0.00 & \textbf{0.092} & 0.05 & \textbf{0.042} \\
        \bottomrule
    \end{tabular}
    \vspace{-0.3cm}
\end{table}

\subsection{RQ1: How does \ourmethod{} perform in multi-agent interaction scenarios?}
\label{sec:crossplay_results}

To answer this question, we conduct cross-play experiments to evaluate \ourmethod{} with prompting-based agents (ReAct, Reflexion, HM) and RL-trained agents (ReAct+RL) across the two environments. \Cref{fig:crossplay_heatmap} presents the results separately for each role, with full numerical results in \cref{tab:cross_play_results}. It can be seen that \ourmethod{} achieves the highest average reward over all opponents in both environments and roles, with gains distributed across most pairings, including interactions with RL-trained opponents. Since model parameters are kept fixed during evaluation, these results demonstrate that the inference and decision capabilities learned through self-play remain effective against agents with different decision frameworks, without additional training for each opponent.

In our environment settings, an invalid action would immediately terminate the episode and incur a \(-3\) penalty for the responsible agent. Therefore, to examine whether the performance gains extend beyond improved compliance with action constraints, we further analyze each agent's per-episode violation rate and its reward on episodes in which neither agent commits a violation, averaged equally over opponents. \cref{tab:validity_diagnostics} first shows that prompting-based agents frequently produce invalid actions in \textit{Kuhn Poker}, while their violation rates remain low in \textit{Bargaining}. This is because the valid actions for agents in \textit{Kuhn Poker} change in different situations, making prompting-based agents easily select invalid actions. In contrast, \ourmethod{} and ReAct+RL exhibit very low violation rates in both environments, indicating that RL training helps agents reliably follow the action constraints. When evaluation is restricted to violation-free episodes, \ourmethod{} still achieves the highest reward over opponents. Its advantage over ReAct+RL, despite their similarly low violation rates, further demonstrates the value of integrating an explicit epistemic layer with action planning.

\begin{figure}[t]
    \centering
    \vspace{-0.3cm}
    \includegraphics[width=0.9\linewidth]{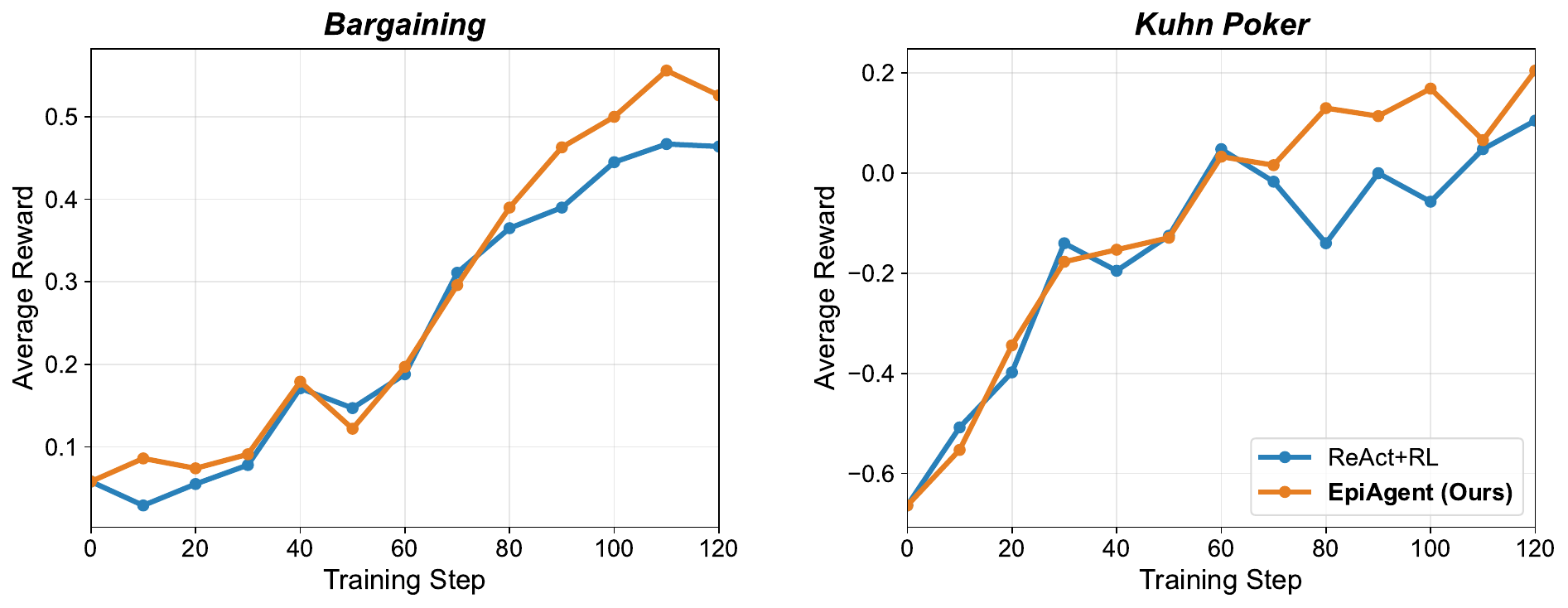}
    \vspace{-0.2cm}
    \caption{Training evaluation of \ourmethod{} versus ReAct+RL baseline (which lacks the epistemic update stage) in both environments. The average reward at each checkpoint is evaluated against a fixed ReAct opponent.}
    \label{fig:training_curves}
    \vspace{-0.2cm}
\end{figure}

\begin{table}[t]
    \centering
    \caption{Ablation results on dedicated epistemic training in both environments. The variant \emph{w/o Epistemic Training} retains both stages but trains only action planning. Each cell reports the average reward as Agent 1 against each opponent, and the \emph{Mean} row is averaged over the five opponents.}
    \label{tab:epistemic_ablation}
    \small
    \renewcommand{\arraystretch}{1.1}
    \begin{tabular}{lcccc}
        \toprule
        \multirow{2}{*}{\textbf{Opponent}} & \multicolumn{2}{c}{\textit{Bargaining}} & \multicolumn{2}{c}{\textit{Kuhn Poker}} \\
        \cmidrule(lr){2-3}\cmidrule(lr){4-5}
        & \multicolumn{1}{c}{\textbf{\ourmethod{}}} & \makecell{\textbf{w/o Epistemic Training}} & \multicolumn{1}{c}{\textbf{\ourmethod{}}} & \makecell{\textbf{w/o Epistemic Training}} \\
        \midrule
        \textbf{ReAct}     & 0.429 & 0.302 & 0.172 & 0.146 \\
        \textbf{Reflexion} & 0.328 & 0.198 & 0.143 & 0.086 \\
        \textbf{HM}        & 0.260 & 0.139 & 0.034 & 0.063 \\
        \textbf{ReAct+RL}  & 0.304 & 0.212 & 0.070 & 0.004 \\
        \textbf{\ourmethod{}} & 0.283 & 0.168 & 0.036 & -0.146 \\
        \midrule
        \rowcolor{cyan!15} \textbf{Mean}  & \textbf{0.321} & 0.204 & \textbf{0.091} & 0.031 \\
        \bottomrule
    \end{tabular}
    \vspace{-0.3cm}
\end{table}

\subsection{RQ2: Does training the epistemic update provide additional performance benefits?}
\label{sec:epistemic_ablation}

To investigate whether training the epistemic update provides additional performance benefits, we compare \ourmethod{} with \textbf{\ourmethod{} w/o Epistemic Training}, which retains both stages but optimizes only action planning using the same RL training procedure. The comparison results are shown in \cref{tab:epistemic_ablation}. \ourmethod{} achieves higher average rewards in both environments, with improvements across most pairings. Since both variants generate beliefs and condition actions on them, these gains demonstrate the benefit of training the epistemic update beyond simply including a belief stage in the decision process. The intrinsic prediction reward provides the epistemic update stage with a dedicated signal on whether its hypotheses about others' private information account for subsequent observations, complementing the environment reward used to train action planning. Learning from this feedback therefore brings additional benefits to the decisions made under the inferred beliefs.

Beyond the final performance gains, we also examine how the explicit epistemic layer and its training affect the whole learning process. \Cref{fig:training_curves} compares \ourmethod{} and ReAct+RL on successive checkpoints against a fixed ReAct opponent, obtained from independent training runs with identical configurations since intermediate checkpoints of the main runs were not retained. Both methods show broadly comparable early progress and improve over training, while \ourmethod{} achieves higher rewards at later checkpoints in both environments. During self-play, evolving opponents continually change the interaction evidence on which the agent bases its decisions. Training epistemic update on this evidence allows EpiAgent to refine its interpretation of others' behavior alongside learning which actions to take, helping belief-conditioned action planning remain responsive to the interactions encountered during training. The performance advantage observed at later checkpoints is consistent with this design, complementing the ablation results and validating the value of coordinating epistemic update and action planning through Epistemic Self-Play.

\begin{table}[t]
    \centering
    \caption{Results of test-time belief interventions, replacing epistemic update outputs (Learned) with Random and Oracle beliefs. Each cell reports the average reward against ReAct using the same trained \ourmethod{} model.}
    \label{tab:belief_intervention}
    \small
    \renewcommand{\arraystretch}{1.1}
    \begin{tabular}{lcccc}
        \toprule
        \multirow{2}{*}{\textbf{Belief Condition}} & \multicolumn{2}{c}{\textit{Bargaining}} & \multicolumn{2}{c}{\textit{Kuhn Poker}} \\
        \cmidrule(lr){2-3} \cmidrule(lr){4-5}
        & \textbf{Agent 0} & \textbf{Agent 1} & \textbf{Agent 0} & \textbf{Agent 1} \\
        \midrule
        \textbf{Random}  & 0.009 & 0.406 & 0.049 & 0.091 \\
        \rowcolor{cyan!15} \textbf{Learned (\ourmethod{})} & 0.160 & 0.429 & 0.181 & 0.172 \\
        \textbf{Oracle}  & 0.197 & 0.506 & 0.232 & 0.195 \\
        \bottomrule
    \end{tabular}
    \vspace{-0.3cm}
\end{table}

\subsection{RQ3: Do learned beliefs provide useful information for action planning?}
\label{sec:belief_results}

\begin{wrapfigure}{r}{0.52\textwidth}
    \centering
    \vspace{-0.3cm}
    \includegraphics[width=0.9\linewidth]{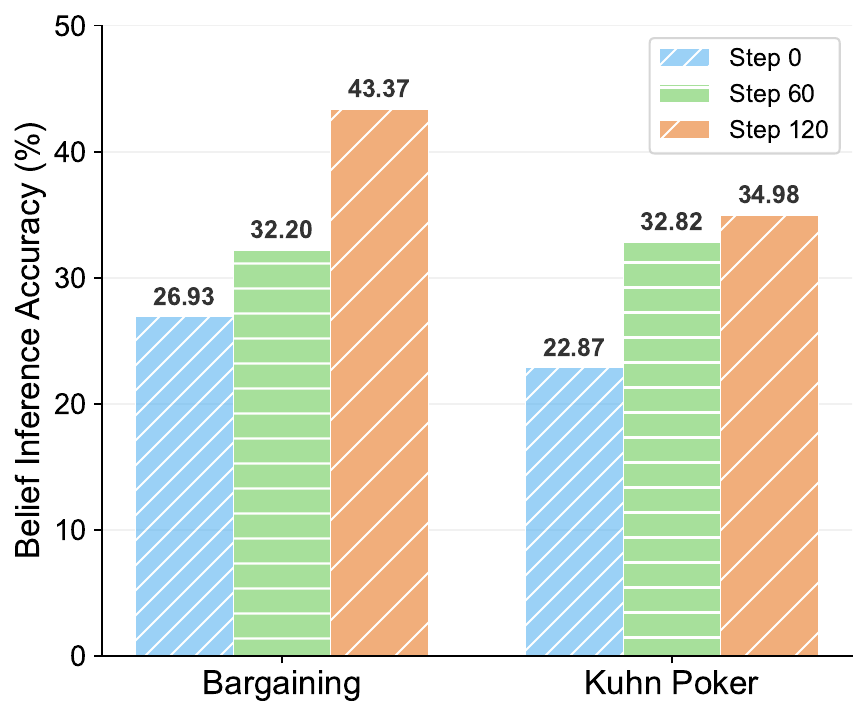}
    \vspace{-0.15cm}
    \caption{Belief inference accuracy of \ourmethod{} at the initial (Step 0), intermediate (Step 60), and final (Step 120) stages of the Epistemic Self-Play training. Each accuracy is evaluated against a fixed ReAct opponent.}
    \label{fig:belief_acc}
    \vspace{-0.2cm}
\end{wrapfigure}

Given that training the epistemic update stage has shown benefits for the agent's decision-making, a natural follow-up question is to what extent Epistemic Self-Play enhances the accuracy of the agent's inferred beliefs. To explore this, we conduct quantitative evaluations of the agent's \textit{belief inference accuracy}, defined as the percentage of times the agent correctly infers the opponent's ground-truth persona, at three training phases: the initial (Step 0), intermediate (Step 60), and final (Step 120) phase against a fixed ReAct opponent. As illustrated in \cref{fig:belief_acc}, belief inference accuracy increases across the evaluated checkpoints in both environments, although ground-truth labels are not used for training. Instead, the intrinsic prediction reward evaluates how well the agent's hypotheses explain subsequent observations. Since personas can influence opponents' behavior, this signal helps the agent identify their private information from behavioral evidence. These results indicate that \ourmethod{}'s gains do not only stem from those training steps over the action planning stage, but are also rooted in a great improvement of the agent's epistemic reasoning.

To further investigate whether these inferred beliefs help action planning, we conduct test-time belief intervention analysis. We evaluate \ourmethod{} under three belief conditions: \textit{Random} samples a persona uniformly and keeps it fixed within each episode, \textit{Learned} uses the model's regular epistemic updates, and \textit{Oracle} supplies the opponent's assigned persona. \Cref{tab:belief_intervention} shows that our method (Learned) obtains higher rewards than Random in both environments and roles, demonstrating the benefit of conditioning action planning on beliefs inferred from the interaction. Since these comparisons use the same trained model, the differences show that its final performance depends on the belief condition supplied to action planning. Oracle further improves rewards in every setting, showing that the planner can exploit access to the opponent's true persona. These results demonstrate that learned beliefs already contribute to effective action planning despite imperfect belief inference, while more accurate beliefs could bring further gains.

\section{Conclusion}
In this work, we present \ourmethod{}, a framework that equips LLM-based agents with an explicit, learnable epistemic layer by decomposing their decision-making process into two coupled stages, \textit{Epistemic Update} and \textit{Action Planning}, that jointly maintain and act on an explicit epistemic state. We formulate their learning dependency as a bi-level optimization problem and propose Epistemic Self-Play, an algorithm that alternately optimizes the two stages through self-play, leveraging an intrinsic reward to train epistemic updates from interaction feedback. Empirical evaluations on \textit{Bargaining} and \textit{Kuhn Poker} demonstrate that \ourmethod{} consistently outperforms both prompting-based and RL-based baselines. Further analyses show that dedicated epistemic training provides additional performance benefits, belief inference improves over training, and learned beliefs contribute to effective action planning. These results demonstrate the value of learning to infer others' private information and act on these inferences, suggesting a path toward LLM-based agents that can adaptively reason about and respond to others in complex social interactions.


\bibliographystyle{unsrtnat}
\bibliography{references}



\appendix

\section{Detailed Environment Settings}
\label[appendix]{appendix:env}
To train and validate our method with other baselines, we construct and modify two distinct yet challenging multi-agent environments: \textit{Bargaining} and \textit{Kuhn Poker}, both of which retain the characteristic of incomplete information and non-stationarity in multi-agent interactions. Specifically, the \textit{Bargaining} combines a shared incentive to reach agreement with conflicting preferences over the deal price: the seller seeks a higher price, while the buyer seeks a lower one. If no agreement is reached, both players receive zero game reward. Therefore, it requires strategic negotiation, concessions, and inference about the opponent's preferences and constraints. Meanwhile, the \textit{Kuhn Poker} is a classic competitive game involving hidden information and deception. Pure strategies in this game are usually bad choices, as they can be exploited easily. So the agent requires the capabilities to make decisions under uncertainty while reasoning about opponents' behavioral tendencies.

\subsection{Bargaining}
\paragraph{Environment Description.}
The \textit{Bargaining} involves two players: a \textbf{Seller} and a \textbf{Buyer}, who negotiate over the deal price of a single item. In this game, both the seller and buyer are assigned private information consisting of a \textit{Target Price}, a \textit{Bottomline Price}, and the item's description (can be different for each player, as the seller usually knows more than the buyer), with an \textit{Initial Price} public to all. All scenario settings in our experiments are extracted from the Craigslist Bargain dataset \citep{he2018decoupling}. The objectives for all players are:
\begin{itemize}[leftmargin=15pt, topsep=3pt, itemsep=0pt]
    \item The \textbf{Seller} aims to \textbf{MAXIMIZE} the final deal price, with its \textit{Target Price} above its \textit{Bottomline Price}. Thus, the seller is incentivized to start negotiations above its target price and concede slowly, strictly rejecting any deal below its bottomline price.

    \item The \textbf{Buyer} aims to \textbf{MINIMIZE} the final deal price, with its \textit{Target Price} below its \textit{Bottomline Price}. So the buyer is supposed to start negotiations below its target price and concede slowly upward, strictly rejecting any deal above its bottomline price.
\end{itemize}
The bargaining process proceeds in a maximum of 10 turns. If both players fail to reach a deal in the end, then this episode will terminate immediately.

\paragraph{Action Space.}
The bargaining process proceeds in turns. Since there is a public initial price, we set the buyer (\textbf{Agent 0}) as the first player in our environment. For the players' action space, in addition to the natural language messages used for negotiation and persuasion, we also employ a set of symbolic actions to formalize the state transitions. These actions are listed as follows:
\begin{itemize}[leftmargin=15pt, topsep=3pt, itemsep=0pt]
    \item \texttt{[OFFER:\$M]}: Make an initial price \$M offer for the item.
    \item \texttt{[COUNTER:\$M]}: Make a counteroffer \$M in response to the opponent's proposal.
    \item \texttt{[QUESTION]}: Ask questions to gather information or clarify details.
    \item \texttt{[REJECT]}: Reject the current offer without making a counteroffer.
    \item \texttt{[ACCEPT]}: Accept the current offer and close the deal.
    \item \texttt{[QUIT]}: End the bargaining without reaching an agreement.
\end{itemize}

\paragraph{Behavioral Personas.}
While LLM-based agents demonstrate strong capabilities in role-playing, they tend to exhibit behavioral homogenization, often converging towards "average" strategies, behavior patterns, or generic preferences present in their pre-training data. This lack of diversity presents two critical issues: (1) It limits the richness of the generated interaction data; (2) It trivializes the challenge of belief inference, as there is little variation in opponents' latent states to distinguish.

To mitigate this and introduce realistic variance into the environment, we employ three distinct yet representative behavioral personas in the \textit{Bargaining} environment:
\begin{itemize}[leftmargin=15pt, topsep=3pt, itemsep=0pt]
    \item \textbf{Patient}: Prioritizes obtaining a favorable price. This persona adopts a "tough" stance, making minimal and slow concessions. They strictly adhere to their target price and are willing to walk away if the offer does not meet their expectations.

    \item \textbf{Balanced}: Balances price with the prospect of reaching an agreement. This persona simulates a rational negotiator who values reciprocity, making gradual concessions in response to the opponent's moves to reach an equitable agreement.

    \item \textbf{Urgent}: Prioritizes reaching a deal quickly over maximizing the price margin. Driven by a preference for reaching agreement quickly, this persona is more likely to make larger and faster concessions to secure an agreement within their bottomline constraints.
\end{itemize}

\paragraph{Reward Design.}
The game reward of each player in the \textit{Bargaining} environment depends on the final deal price and their role. If no deal is reached, both players receive zero game reward. Given a deal at terminal step \(T\) with price \(p^\star\), the \textbf{Seller} and \textbf{Buyer}'s rewards are defined as:
\begin{equation}
    r_\text{S}^{T} = \frac{p^\star - p_\text{S}^\text{bottom}}{p_\text{S}^\text{target} - p_\text{S}^\text{bottom}}, \quad r_\text{B}^{T} = \frac{p_\text{B}^\text{bottom} - p^\star}{p_\text{B}^\text{bottom} - p_\text{B}^\text{target}},
\end{equation}
where \(p_\text{S}^\text{bottom}\) and \(p_\text{S}^\text{target}\) denote the seller's bottomline and target price, respectively, with analogous definitions for the buyer. For each player, the reward equals zero at its bottomline and one at its target, and lies in \([0,1]\) when the deal price falls between these thresholds. Rewards are not clipped: outcomes more favorable than the target yield rewards above one, while outcomes worse than the bottomline yield negative rewards.

Besides the game reward, we also introduce a format penalty (\(-3\)) in the textual environment to ensure parsable outputs from the LLM-based agents. And once an invalid response is taken, the environment will terminate immediately.

\subsection{Kuhn Poker}
\paragraph{Environment Description.}
The \textit{Kuhn Poker} is a simplified form of poker, serving as a simple model of a zero-sum, two-player, imperfect information game. In this game, the deck includes only three playing cards, usually a Jack, Queen, and King. Both players are dealt a card, and their available actions depend on the current betting state. If both players check, or a bet is called, the player with the higher card wins at showdown. If a player folds, the opponent wins the pot. The opponent's card is revealed only at showdown.

One round of the \textit{Kuhn Poker} has a maximum depth of 3 actions. The dynamics allow for complex strategic maneuvers, such as bluffing (betting with a weak card) or slow-play (checking with a strong card to induce a bet). To better help the agent infer the opponent's potential strategy style, we extend the gameplay to 5 rounds with newly dealt cards in each round, which means there are a maximum of 15 turns in an interaction episode. And we set player 1 (\textbf{Agent 1}) as the first player in this environment.

\paragraph{Action Space.}
Different from the \textit{Bargaining} game, where the player's action contains natural language messages, the action space in the \textit{Kuhn Poker} game is discrete and only contains four possible actions:
\begin{itemize}[leftmargin=15pt, topsep=3pt, itemsep=0pt]
    \item \texttt{[check]}: Pass without betting (only if no bet is on the table).
    \item \texttt{[bet]}: Add 1 chip to the pot (only if no bet is on the table).
    \item \texttt{[call]}: Match an opponent's bet by adding 1 chip to the pot.
    \item \texttt{[fold]}: Surrender and let your opponent win the pot.
\end{itemize}

\paragraph{Behavioral Personas.}
Similarly, to increase the diversity of the interaction trajectories and align more closely with the strategic biases of human players, we also employ four representative behavioral personas in the \textit{Kuhn Poker} environment. Each player is privately assigned a persona through its prompt, and this assignment remains fixed across all five rounds:
\begin{itemize}[leftmargin=15pt, topsep=3pt, itemsep=0pt]
    \item \textbf{Balanced (GTO)}: Approximates a Game Theoretic Optimal strategy. This persona encourages mixed strategies, such as probabilistic bluffing and calling, to maintain unpredictable behavior across hands.

    \item \textbf{Aggressive (Over-Bluffer)}: Adopts a hyper-offensive style. This persona bets frequently with weak hands (J) to pressure opponents, relying on intimidation rather than card strength.

    \item \textbf{Conservative (Folder)}: Exhibits extreme risk aversion. This persona folds frequently when facing aggression and only commits chips to the pot when holding strong cards (K), making them easy to bluff but dangerous to call.

    \item \textbf{Calling Station (Over-Caller)}: A passive but stubborn style. This persona rarely folds and calls excessively, operating under the belief that opponents are constantly bluffing. They maximize showdowns regardless of expected value.
\end{itemize}

\paragraph{Reward Design.}
The game reward of each player in the \textit{Kuhn Poker} environment only depends on their final chips after 5 rounds. The player with more chips wins and receives \(1\) reward, while its opponent receives \(-1\) reward. If both players have the same chips (\textit{i.e.}, they tie), then both of them receive \(0\) reward.

Also, we add a penalty of \(-3\) for invalid responses or actions, including parsing failures and actions that are illegal in the current betting state. The penalty applies to the responsible player, and the environment will terminate the episode immediately if it receives such a response.

\begin{table*}[t]
    \centering
    \vspace{-0.3cm}
    \caption{Cross-play evaluation results in \textit{Bargaining} and \textit{Kuhn Poker}. Each cell reports the average reward (with format penalty) and standard deviation for a specific Agent 0 (row) v.s. Agent 1 (column) pairing: the upper number is Agent 0's average reward, and the lower one is Agent 1's. \textcolor{red}{Red} values mark the best Agent 0 per column; the \textcolor{blue}{blue} values mark the best Agent 1 per row.}
    \vspace{0.3cm}
    \label{tab:cross_play_results}
    \renewcommand{\arraystretch}{1.3}
    \begin{tabular}{lccccc}
        \toprule
        \multirow{2}{*}{\textbf{Agent 0}} & \multicolumn{5}{c}{\textbf{Agent 1}} \\
        \cmidrule(lr){2-6}
         & \textbf{ReAct} & \textbf{Reflexion} & \textbf{HM} & \textbf{ReAct+RL} & \textbf{\ourmethod{}} \\
        \midrule
        \multicolumn{6}{c}{\textit{Bargaining}} \\
        \midrule
        \textbf{ReAct} & \makecell[rt]{\meanstd{-0.052}{0.021} \\ \meanstd{0.326}{0.009}} & \makecell[rt]{\meanstd{-0.131}{0.086} \\ \meanstd{0.295}{0.047}} & \makecell[rt]{\meanstd{-0.236}{0.219} \\ \meanstd{0.329}{0.129}} & \makecell[rt]{\meanstd{-0.155}{0.038} \\ \meanstd{0.390}{0.045}} & \makecell[rt]{\meanstd{-0.136}{0.043} \\ \textcolor{blue}{\meanstd{0.429}{0.031}}} \\
        \textbf{Reflexion} & \makecell[rt]{\meanstd{0.017}{0.015} \\ \meanstd{0.187}{0.028}} & \makecell[rt]{\meanstd{-0.005}{0.045} \\ \meanstd{0.027}{0.055}} & \makecell[rt]{\meanstd{-0.136}{0.161} \\ \meanstd{0.164}{0.133}} & \makecell[rt]{\meanstd{-0.018}{0.135} \\ \meanstd{0.218}{0.064}} & \makecell[rt]{\meanstd{-0.020}{0.056} \\ \textcolor{blue}{\meanstd{0.328}{0.022}}} \\
        \textbf{HM} & \makecell[rt]{\meanstd{-0.055}{0.119} \\ \meanstd{0.096}{0.082}} & \makecell[rt]{\meanstd{-0.229}{0.094} \\ \meanstd{0.070}{0.087}} & \makecell[rt]{\meanstd{-0.332}{0.047} \\ \meanstd{0.206}{0.053}} & \makecell[rt]{\meanstd{-0.214}{0.101} \\ \meanstd{0.133}{0.120}} & \makecell[rt]{\meanstd{-0.182}{0.220} \\ \textcolor{blue}{\meanstd{0.260}{0.147}}} \\
        \textbf{ReAct+RL} & \makecell[rt]{\meanstd{0.104}{0.044} \\ \meanstd{0.175}{0.015}} & \makecell[rt]{\meanstd{0.050}{0.050} \\ \meanstd{0.147}{0.043}} & \makecell[rt]{\meanstd{-0.014}{0.131} \\ \meanstd{0.177}{0.074}} & \makecell[rt]{\meanstd{0.023}{0.008} \\ \meanstd{0.238}{0.015}} & \makecell[rt]{\meanstd{0.041}{0.033} \\ \textcolor{blue}{\meanstd{0.304}{0.025}}} \\
        \textbf{\ourmethod{}} & \makecell[rt]{\textcolor{red}{\meanstd{0.160}{0.020}} \\ \meanstd{0.227}{0.017}} & \makecell[rt]{\textcolor{red}{\meanstd{0.108}{0.055}} \\ \meanstd{0.239}{0.047}} & \makecell[rt]{\textcolor{red}{\meanstd{0.014}{0.020}} \\ \textcolor{blue}{\meanstd{0.298}{0.026}}} & \makecell[rt]{\textcolor{red}{\meanstd{0.091}{0.069}} \\ \meanstd{0.261}{0.034}} & \makecell[rt]{\textcolor{red}{\meanstd{0.085}{0.031}} \\ \meanstd{0.283}{0.029}} \\
        \midrule
        \multicolumn{6}{c}{\textit{Kuhn Poker}} \\
        \midrule
        \textbf{ReAct} & \makecell[rt]{\meanstd{-0.786}{0.094} \\ \meanstd{-0.151}{0.138}} & \makecell[rt]{\meanstd{-0.734}{0.035} \\ \meanstd{-1.029}{0.043}} & \makecell[rt]{\meanstd{-0.797}{0.081} \\ \meanstd{-0.760}{0.183}} & \makecell[rt]{\meanstd{-0.792}{0.030} \\ \textcolor{blue}{\meanstd{0.190}{0.037}}} & \makecell[rt]{\meanstd{-0.898}{0.075} \\ \meanstd{0.172}{0.090}} \\
        \textbf{Reflexion} & \makecell[rt]{\meanstd{-1.425}{0.072} \\ \meanstd{-0.263}{0.073}} & \makecell[rt]{\meanstd{-1.170}{0.076} \\ \meanstd{-0.802}{0.190}} & \makecell[rt]{\meanstd{-1.424}{0.152} \\ \meanstd{-0.638}{0.139}} & \makecell[rt]{\meanstd{-1.695}{0.041} \\ \meanstd{0.125}{0.023}} & \makecell[rt]{\meanstd{-1.714}{0.099} \\ \textcolor{blue}{\meanstd{0.143}{0.086}}} \\
        \textbf{HM} & \makecell[rt]{\meanstd{-1.714}{0.151} \\ \meanstd{-0.242}{0.082}} & \makecell[rt]{\meanstd{-1.349}{0.195} \\ \meanstd{-1.266}{0.219}} & \makecell[rt]{\meanstd{-1.587}{0.090} \\ \meanstd{-0.711}{0.068}} & \makecell[rt]{\meanstd{-1.708}{0.047} \\ \meanstd{-0.010}{0.030}} & \makecell[rt]{\meanstd{-1.940}{0.114} \\ \textcolor{blue}{\meanstd{0.034}{0.033}}} \\
        \textbf{ReAct+RL} & \makecell[rt]{\meanstd{0.109}{0.035} \\ \meanstd{-0.461}{0.075}} & \makecell[rt]{\meanstd{-0.026}{0.058} \\ \meanstd{-1.891}{0.077}} & \makecell[rt]{\meanstd{-0.060}{0.061} \\ \meanstd{-1.143}{0.091}} & \makecell[rt]{\meanstd{0.000}{0.014} \\ \meanstd{-0.008}{0.000}} & \makecell[rt]{\meanstd{-0.078}{0.048} \\ \textcolor{blue}{\meanstd{0.070}{0.041}}} \\
        \textbf{\ourmethod{}} & \makecell[rt]{\textcolor{red}{\meanstd{0.181}{0.076}} \\ \meanstd{-0.445}{0.051}} & \makecell[rt]{\textcolor{red}{\meanstd{0.019}{0.066}} \\ \meanstd{-2.003}{0.107}} & \makecell[rt]{\textcolor{red}{\meanstd{-0.052}{0.055}} \\ \meanstd{-1.260}{0.063}} & \makecell[rt]{\textcolor{red}{\meanstd{0.096}{0.145}} \\ \meanstd{-0.096}{0.145}} & \makecell[rt]{\textcolor{red}{\meanstd{-0.044}{0.115}} \\ \textcolor{blue}{\meanstd{0.036}{0.107}}} \\
        \bottomrule
    \end{tabular}
    \vspace{-0.3cm}
\end{table*}

\section{Additional Experiment Results}
\label[appendix]{appendix:add_exp}

\subsection{Full Cross-Play Rewards}
\label{appendix:crossplay_rewards}
\Cref{tab:cross_play_results} provides the numerical rewards underlying the main heatmap, including the reported standard deviations. The table preserves the original pairing layout: rows identify Agent 0 and columns identify Agent 1, with both agents' rewards listed in each cell.

\subsection{Ablation Results of Switch Frequency}
\label[appendix]{appendix:freq}
We conducted an additional ablation study on \textit{Kuhn Poker} with different switch frequencies \(L \in \{5, 10, 15\}\). After training, we select the checkpoint with the best combined performance of both agents during self-play training. This criterion relies only on training-time self-play statistics and does not involve any evaluation opponent. The selected checkpoints are then evaluated against a fixed ReAct opponent, and the average rewards are reported in \cref{tab:switch_freq}.

\begin{table}[!ht]
    \centering
    \caption{Performance of \ourmethod{} under different training switch frequencies \(L\).}
    \label{tab:switch_freq}
    \renewcommand{\arraystretch}{1.1}
    \begin{tabular}{cccc}
        \toprule
        \textbf{Switch Frequency \(L\)} & \textbf{5} & \textbf{10} & \textbf{15} \\
        \midrule
        \textbf{Average Reward} & 0.041 & 0.181 & 0.039 \\
        \bottomrule
    \end{tabular}
    \vspace{-0.1cm}
\end{table}

The results show that both \(L = 5\) and \(L = 15\) yield lower performance than \(L = 10\), which is consistent with our intuition: switching too frequently (small \(L\)) may leave insufficient time to improve each stage under its current objective, while switching too infrequently (large \(L\)) may delay each stage's response to changes in the other. This suggests that the switch frequency should be chosen to balance these two extremes, and \(L = 10\) serves as an effective choice in our setting.

\section{Detailed Experiment Settings}
\label[appendix]{appendix:exp_details}

\subsection{Training Details}
We implement the training pipeline based on RAGEN \citep{wang2025ragen} and VeRL \citep{sheng2025hybridflow}. As the original RAGEN framework only supports training on a single agent, we modified it to be suitable for our multi-agent rollout and training. Due to resource constraints, we select REINFORCE++ \citep{hu2025reinforce} as the RL training algorithm to save resources and for efficient training. We employ Qwen2.5-3B-Instruct as our base model for all experiments and use LoRA for the model training.

During the multi-agent rollout, we use the same LLM to perform self-play to collect interaction trajectories. As our experimental environments are multi-turn, multi-agent interaction scenarios, each agent acts as the "assistant" in its own message list, while the environment feedback and other agents' actions act as the "user". For the rollout hyperparameters, we set temperature=1.0, top\_p=1.0, top\_k=-1, and use vLLM for inference acceleration. Considering the randomness of LLM's generation, at each training step, we initialize 32 environment groups with different random seeds, each containing 16 environments with identical initialization, and let the LLM agents perform rollouts across all \(32 \times 16 = 512\) environments in parallel. This yields 512 trajectories per agent. Following the rollout filtering technique in RAGEN, we adopt the within-group standard deviation of returns as the filtering criterion and set filter\_ratio=0.25, retaining 128 trajectories per agent after filtering. Finally, since we mix trajectories from all agents during training, each training step uses \(128 \times 2 = 256\) trajectories under our two-agent settings.

During training, we set the LoRA rank to 32 and alpha to 16, with LoRA adapters applied to all linear layers of the base model. Since the optimal learning rate for LoRA training usually differs from that of full fine-tuning, we adopt the Adam optimizer with a learning rate of 7e-6. Following common practice in LLM training, we also apply a KL-divergence constraint between the current policy and the reference model, using the K2 estimator with a coefficient of 0.001. Based on the ablation results in \cref{appendix:freq}, we set the switch frequency between the two stages of Epistemic Self-Play to 10, which achieves the highest average reward among the tested settings. All other training-related hyperparameters follow the default settings of RAGEN and VeRL.

\subsection{Computational Consumption}
\label[appendix]{appendix:consumption}
All experiments are conducted on a node equipped with 8 NVIDIA A100 GPUs. We use bfloat16, Flash-Attention, and gradient checkpointing to run our experiments. For each environment, a 120-step training run takes approximately 40 hours, with most of the time spent on multi-agent rollouts.

\section{Prompt Templates}
\label[appendix]{appendix:prompt}
In this section, we present all prompt templates used in our experiments. Specifically, the epistemic update and action planning prompts are used for \ourmethod{}'s decision-making process, while the opponent prediction prompt is leveraged to construct the belief-conditioned prediction trajectories, which are further used to compute the agent's intrinsic prediction reward based on the LLM's output log-probabilities.

\subsection{Bargaining}
\begin{tcolorbox}[title=\textbf{Environment Prompt}, breakable, fontupper=\ttfamily, left=3mm, right=3mm]
You are participating in a bargaining negotiation between a seller and a buyer over an item. You will be assigned either the seller or buyer role.
\\

\#\# Your Capabilities

You can perform multiple negotiation-related tasks:

- **Belief Inference**: Analyze opponent behavior to infer their strategy and type

- **Action Planning**: Determine optimal negotiation actions and messages

- **Opponent Prediction**: Predict opponent's most likely next moves and responses
\\

\#\# Game Framework
\\

\#\#\# Roles and Objectives

- **Seller**: Maximize sale price within constraints

- **Buyer**: Minimize purchase price within constraints

- Each player has private information (target price, bottomline price, type, etc.)
\\

\#\#\# Price Parameters

- **Target Price**: Your ideal outcome that balances ambition with realism - the PRIMARY objective to aim for \\
    - Sellers: Start ABOVE your target price, concede slowly downward \\
    - Buyers: Start BELOW your target price, concede slowly upward \\
- **Bottomline Price**: Your walk-away point - crossing this means no deal is better than a bad deal \\
    - Seller: The minimum price you will accept (never go below this) \\
    - Buyer: The maximum price you will pay (never go above this) \\

\#\#\# Available Actions

- **OFFER: \$M**: Make an initial price \$M offer for the item

- **COUNTER: \$M**: Make a counteroffer \$M in response to the opponent's proposal

- **QUESTION**: Ask questions to gather information or clarify details

- **REJECT**: Reject the current offer without making a counteroffer

- **ACCEPT**: Accept the current offer and close the deal

- **QUIT**: End the negotiation without reaching an agreement
\\

\#\#\# Player Types

Participants exhibit three main types:

- **Patient**: Prioritize maximizing own benefit, make minimal and slow concessions, willing to walk away if the price is not close to the target, rarely compromise

- **Balanced**: Seek fair outcomes with moderate patience, make gradual and reciprocal concessions, willing to negotiate but won't accept unfair deals

- **Urgent**: More likely to make or accept larger and faster concessions to reach agreement, prefer securing a deal over maximizing price
\\

\#\#\# Core Principles

- **Maximize Your Outcome**: ALWAYS aim for the best possible price towards your target

- **Strategic Communication**: Justify offers with concrete reasons (quality, features, market value); use persuasive framing to influence opponent

- **Strategic Concessions**: Make small, justified concessions strategically, not eagerly

- **Effective Anchoring**: Initial offers shape negotiation range; start with room for strategic concessions toward target

- **Bottomline is Sacred**: NEVER cross your bottomline under any circumstances

- **Valid Actions**: Only use OFFER, COUNTER, QUESTION, REJECT, ACCEPT, or QUIT
\\

**Remember**: You are a strategic agent with preferences, not a random actor or a rigid rule-follower. Make decisions that serve your goals while adapting to the strategic context. And always output in English.
\end{tcolorbox}

\begin{tcolorbox}[title=\textbf{Epistemic Update Prompt}, breakable, fontupper=\ttfamily, left=3mm, right=3mm]
State:

Now it is Turn \{TURN\_NUM\} in the negotiation. You are the **\{ROLE\}** during the bargaining negotiation.
\\

\#\# Product Information

- Name: \{PRODUCT\_NAME\}

- Description: \{PRODUCT\_DESC\}

- Initial Price: \$\{INITIAL\_PRICE\}
\\

\#\# Your Negotiation Type

\{PLAYER\_TYPE\_DESC\}
\\

\#\# Negotiation Process

- Recent Messages:

\{RECENT\_MESSAGES\}
\\

- Offer History:

\{OFFER\_HISTORY\}
\\

\#\# Task: Infer Your Opponent's Type

Based on:

- The negotiation history and opponent's behavioral patterns

- Your observations of their concession style, time sensitivity, and risk tolerance

- Your own type's perspective on their actions

Analyze your opponent (\{OPPONENT\_ROLE\})'s behavior to estimate THEIR type, target price, and bottomline.
\\

\#\# Output Format

Always output: `\{RESPONSE\_FORMAT\}` with no extra text.

The inferred belief about your opponent must be enclosed within the <answer></answer> tag with the structure format:

`<answer>**ESTIMATED\_OPPONENT\_TYPE**: [Patient / Balanced / Urgent] (Confidence: [High / Medium / Low])

**ESTIMATED\_OPPONENT\_TARGET**: \$X (Range: \$Y - \$Z)

**ESTIMATED\_OPPONENT\_BOTTOMLINE**: \$A (Range: \$B - \$C)

**STRATEGIC\_ASSESSMENT**: [1-2 sentences summarizing their overall strategy]</answer>`
\\

You have \{TURNS\_LEFT\} turns left. Strictly follow the output format. Max response length: \{MAX\_TOKENS\} tokens.
\end{tcolorbox}

\begin{tcolorbox}[title=\textbf{Action Planning Prompt}, breakable, fontupper=\ttfamily, left=3mm, right=3mm]
State:

Now it is Turn \{TURN\_NUM\} in the negotiation. You are the **\{ROLE\}** during the bargaining negotiation.
\\

\#\# Product Information

- Name: \{PRODUCT\_NAME\}

- Description: \{PRODUCT\_DESC\}

- Initial Price: \$\{INITIAL\_PRICE\}
\\

\#\# Your Private Information

- Your Target: \$\{TARGET\_PRICE\}

- Your Bottomline: \$\{BOTTOMLINE\_PRICE\}
\\

\#\# Your Negotiation Type

\{PLAYER\_TYPE\_DESC\}
\\

\#\# Negotiation Process

- Recent Messages:

\{RECENT\_MESSAGES\}
\\

- Offer History:

\{OFFER\_HISTORY\}
\\

\#\# Your Latest Belief About Opponent

\{BELIEF\}
\\

\#\# Task: Plan Your Next Action

Based on:

- The negotiation history and current offer on the table

- Your latest belief about the opponent's type and price range

- Your own type's values and strategic priorities

- Remaining turns and negotiation phase

Determine your next negotiation move strategically.

**Remember**: Align with your type's values while remaining strategically adaptive.
\\

\#\# Output Format

Always output: `\{RESPONSE\_FORMAT\}` with no extra text.

All responses enclosed within the <answer></answer> tag must follow this format with NO exceptions: `<answer>[ACTION] your negotiation message</answer>`

The ACTION MUST be enclosed in [square brackets], and the message MUST be a complete sentence and align with the ACTION.
\\

\#\#\# Output Example

You are a buyer with target price \$50 and bottomline price \$80. The seller offers \$100 for an item.

Example responses:

- `<answer>[COUNTER: \$60] I think \$100 is too high. How about \$60?</answer>`

- `<answer>[QUESTION] What's the condition of the laptop? Any warranty included?</answer>`

- `<answer>[REJECT] That price is way beyond my budget.</answer>`
\\

You have \{TURNS\_LEFT\} turns left. Strictly follow the output format. Max response length: \{MAX\_TOKENS\} tokens.
\end{tcolorbox}

\begin{tcolorbox}[title=\textbf{Opponent Prediction Prompt}, breakable, fontupper=\ttfamily, left=3mm, right=3mm]
State:

Now it is Turn \{TURN\_NUM\} in the negotiation. You are the **\{ROLE\}** during the bargaining negotiation.
\\

\#\# Product Information

- Name: \{PRODUCT\_NAME\}

- Description: \{PRODUCT\_DESC\}

- Initial Price: \$\{INITIAL\_PRICE\}
\\

\#\# Your Negotiation Type

\{PLAYER\_TYPE\_DESC\}
\\

\#\# Negotiation Process

- Recent Messages:

\{RECENT\_MESSAGES\}
\\

- Offer History:

\{OFFER\_HISTORY\}
\\

\#\# Your Latest Belief About Opponent

\{BELIEF\}
\\

\#\# Your Proposed Action

\{ACTION\}
\\

\#\# Task: Predict Your Opponent's Response

Based on:

- Your belief about the opponent's type, target, and bottomline

- The negotiation history and their behavioral pattern

- Your proposed action above

Predict your opponent's most likely response to your proposed action.

**Remember**: Think from the opponent's perspective, following their expected negotiation pattern. Output ONLY the opponent's predicted response, NO extra content.
\\

\#\# Output Format

Always output: `\{RESPONSE\_FORMAT\}` with no extra text.

Your predicted response enclosed within the <answer></answer> tag must follow this format with NO exceptions: `<answer>OPPONENT\_ROLE: their predicted message</answer>`
\\

\#\#\# Output Example

You are a buyer who believes the opponent is a **Patient** seller (estimated target: \$80, bottomline: \$60). You just proposed: 'Given the condition, I think \$40 is fair.'

Example prediction:

- `<answer>seller: That's too low for an almost brand-new item. I can come down to \$75, but that's as low as I'm willing to go.</answer>`
\\

You have \{TURNS\_LEFT\} turns left. Strictly follow the output format. Max response length: \{MAX\_TOKENS\} tokens.
\end{tcolorbox}

\subsection{Kuhn Poker}
\begin{tcolorbox}[title=\textbf{Environment Prompt}, breakable, fontupper=\ttfamily, left=3mm, right=3mm]
You are playing Kuhn Poker, a simplified two-player poker game. Make valid actions to win.
\\

\#\# Your Capabilities

You can perform multiple game-related tasks:

- **Belief Inference**: Analyze opponent behavior to infer their types and strategy styles

- **Action Planning**: Determine optimal actions based on your beliefs about the opponent

- **Opponent Prediction**: Predict opponent's most likely next moves based on the interaction histories and your beliefs
\\

\#\# Game Framework
\\

\#\#\# Game Rules

- Kuhn Poker uses a 3-card deck with J, Q, K (J lowest, K highest)

- Each player antes 1 chip and receives 1 card each round (note that the cards are dealt without replacement, so you cannot have the same card as your opponent)

- Game continues for 5 rounds

- The player with the **most chips after all rounds** wins
\\

\#\#\# Action Rules

- '[check]': Pass without betting (only if no bet is on the table)

- '[bet]': Add 1 chip to the pot (only if no bet is on the table)

- '[call]': Match an opponent's bet by adding 1 chip to the pot

- '[fold]': Surrender your hand and let your opponent win the pot
\\

\#\#\# Player Types

Players in Kuhn Poker exhibit four main strategy styles:

- **Balanced (GTO)**: Play near-optimal mixed strategies to remain unexploitable, use theoretically sound bluffing and calling frequencies, maintain unpredictable patterns across hands

- **Aggressive (Over-Bluffer)**: Bet and bluff frequently with weak hands to pressure opponents into folding, highly offensive and willing to take risks, prefer winning through intimidation

- **Conservative (Folder)**: Extremely risk-averse and cautious, fold frequently when facing bets, rarely bluff with weak hands, only commit chips with strong confidence

- **Calling Station (Over-Caller)**: Call excessively and rarely fold, reluctant to give up on any hand, believe opponents are often bluffing, prefer seeing showdowns over folding
\end{tcolorbox}

\begin{tcolorbox}[title=\textbf{Epistemic Update Prompt}, breakable, fontupper=\ttfamily, left=3mm, right=3mm]
State:

\{GAME\_OBSERVATION\}
\\

\#\# Task: Infer Your Opponent's Type

Based on:

- The interaction history and opponent's behavioral patterns

- Your observations of their strategy style

- Your own type's perspective on their actions

Analyze your opponent's behavior to estimate THEIR type and strategy.
\\

\#\# Output Format

Always output: `\{RESPONSE\_FORMAT\}` with no extra text.

The inferred belief about your opponent must be enclosed within the <answer></answer> tag with the structure format:

`<answer>**ESTIMATED\_OPPONENT\_TYPE**: [Balanced / Aggressive / Conservative / Calling Station] (Confidence: [High / Medium / Low])

**STRATEGIC\_ASSESSMENT**: [1-2 sentences summarizing their overall strategy and tendencies]</answer>`
\\

Strictly follow the output format. Max response length: \{MAX\_TOKENS\} tokens.
\end{tcolorbox}

\begin{tcolorbox}[title=\textbf{Action Planning Prompt}, breakable, fontupper=\ttfamily, left=3mm, right=3mm]
State:

\{GAME\_OBSERVATION\}
\\

\#\# Your Latest Belief About Opponent

\{BELIEF\}
\\

\#\# Task: Plan Your Next Action

Based on:

- The interaction history and current observation

- Your latest belief about the opponent's type and strategy style

- Your own type's values and strategic priorities

Determine your next move strategically.

**Remember**: Align with your type's values while remaining strategically adaptive.
\\

\#\# Output Format

Always output: `\{RESPONSE\_FORMAT\}` with no extra text.

All responses enclosed within the <answer></answer> tag must follow this format with NO exceptions: `<answer>[ACTION]</answer>`

The ACTION MUST be enclosed in [square brackets], and you MUST choose exactly one action from the current available actions above.
\\

Strictly follow the output format. Max response length: \{MAX\_TOKENS\} tokens.
\end{tcolorbox}

\begin{tcolorbox}[title=\textbf{Opponent Prediction Prompt}, breakable, fontupper=\ttfamily, left=3mm, right=3mm]
State:

\{GAME\_OBSERVATION\}
\\

\#\# Your Chosen Action

\{ACTION\}
\\

\#\# Your Latest Belief About Opponent

\{BELIEF\}
\\

\#\# Task: Predict Your Opponent's Move

Based on:

- Your belief about the opponent's type and strategy style

- The interaction history and their behavioral pattern

- Your chosen action above

Predict your opponent's most likely move to your chosen action.

**Remember**: Think from the opponent's perspective, following their expected behavioral pattern. Output ONLY the opponent's predicted move, NO extra content.
\\

\#\# Output Format

Always output: `\{RESPONSE\_FORMAT\}` with no extra text.

Your predicted move enclosed within the <answer></answer> tag must follow this format with NO exceptions: `<answer>[PREDICTED\_OPPONENT\_ACTION]</answer>`
\\

Strictly follow the output format. Max response length: \{MAX\_TOKENS\} tokens.
\end{tcolorbox}

\end{document}